%% file: acmmm-paper.tex
\pdfoutput=1

\documentclass[sigconf]{acmart}

\usepackage{amsmath}
\usepackage{xspace}
\usepackage{algorithm}
\usepackage[noend]{algpseudocode}
\usepackage{tabularx}
\usepackage{array}
\usepackage{calc}
\usepackage{booktabs}
\usepackage{pgfplots}
\usepackage{sansmath}
\usepackage{standalone}
\usepackage{threeparttable}
\usepackage{cleveref}
\usepackage[justification=centering, caption=false]{subfig}

\setcopyright{none}

\usetikzlibrary{shapes}
\usetikzlibrary{positioning}
\usetikzlibrary{matrix}
\usetikzlibrary{calc}
\usetikzlibrary{decorations.pathreplacing}

\input{fig/quantizer/cmaps}
\pgfplotsset{compat=1.12}

\newcommand*{\eg}{e.g.,\xspace}
\newcommand*{\ie}{i.e.,\xspace}
\newcommand*{\etc}{etc.\xspace}

\DeclareMathOperator*{\argmin}{arg\,min}

\newcommand*{\stimes}{{\times}}
\newcommand*{\pq}[2]{PQ\,${#1}\stimes{#2}$}
\newcommand*{\opq}[2]{OPQ\,${#1}\stimes{#2}$}



\crefformat{footnote}{#2\footnotemark[#1]#3}

\begin{document}

\title{Derived Codebooks for High-Accuracy Nearest Neighbor Search}

\author{Fabien Andr\'e}
\affiliation{
	\institution{Technicolor}
}

\author{Anne-Marie Kermarrec}
\affiliation{
	\institution{Inria}
}

\author{Nicolas Le Scouarnec}
\affiliation{
	\institution{Technicolor}
}

\copyrightyear{2019}
\acmYear{2019}
\setcopyright{none}
\acmConference{}{}{}
\acmPrice{}
\acmDOI{}
\acmISBN{}

\settopmatter{printacmref=false}
\setcopyright{none}
\renewcommand\footnotetextcopyrightpermission[1]{} 
\pagestyle{plain} 

\begin{abstract}
\input{abstract.tex}

\end{abstract}

\maketitle

\section{Introduction}
\input{introduction.tex}

\section{Background}
\input{background.tex}
\section{Derived Codebooks}
\input{joint_quantizers.tex}

\section{Evaluation}
\input{evaluation.tex}

\section{Discussion}
\input{related.tex}

\section{Conclusion}
\input{conclusion.tex}


%
\bibliographystyle{ACM-Reference-Format}
\bibliography{acmmm-paper}
\end{document}

%% file: fig/quantizer/cmaps.tex
\pgfplotsset{
    /pgfplots/colormap={jet}{rgb255(0cm)=(0,0,128) rgb255(1cm)=(0,0,255)
    rgb255(3cm)=(0,255,255) rgb255(5cm)=(255,255,0) rgb255(7cm)=(255,0,0) rgb255(8cm)=(128,0,0)}
}

\pgfplotsset{
    colormap={bluered}{
        rgb255(0cm)=(0,0,180); rgb255(1cm)=(0,255,255); rgb255(2cm)=(100,255,0);
        rgb255(3cm)=(255,255,0); rgb255(4cm)=(255,0,0); rgb255(5cm)=(128,0,0)}
}

\pgfplotsset{
    colormap={cubehelix16}{
        rgb255(0cm)=(0,0,0);
        rgb255(1cm)=(22,10,34);
        rgb255(2cm)=(24,32,68);
        rgb255(3cm)=(16,62,83);
        rgb255(4cm)=(14,94,74);
        rgb255(5cm)=(35,116,51);
        rgb255(6cm)=(80,125,35);
        rgb255(7cm)=(138,122,45);
        rgb255(8cm)=(190,117,85);
        rgb255(9cm)=(218,121,145);
        rgb255(10cm)=(219,138,203);
        rgb255(11cm)=(204,167,240);
        rgb255(12cm)=(191,201,251);
        rgb255(13cm)=(195,229,244);
        rgb255(14cm)=(220,246,239);
        rgb255(15cm)=(255,255,255)}
}

\pgfplotsset{
    colormap={rainbow16}{
        rgb255(0cm)=(135,59,97);
        rgb255(1cm)=(143,64,127);
        rgb255(2cm)=(143,72,157);
        rgb255(3cm)=(135,85,185);
        rgb255(4cm)=(121,102,207);
        rgb255(5cm)=(103,123,220);
        rgb255(6cm)=(84,146,223);
        rgb255(7cm)=(69,170,215);
        rgb255(8cm)=(59,192,197);
        rgb255(9cm)=(60,210,172);
        rgb255(10cm)=(71,223,145);
        rgb255(11cm)=(93,229,120);
        rgb255(12cm)=(124,231,103);
        rgb255(13cm)=(161,227,95);
        rgb255(14cm)=(198,220,100);
        rgb255(15cm)=(233,213,117)
    }
}

\tikzset{adjust near node color/.code={%
    \pgfplotscolormapdefinemappedcolor\pgfplotspointmetatransformed%
    \definecolor{mapped node color}{rgb}{\pgfmathresult}%
    \pgfkeys{/tikz/text=mapped node color!80!black}%
    }
}

\pgfplotsset{voronoiaxis/.style={restrict x to domain*=-5:5,
    restrict y to domain*=-5:5,
    xmin=0,
    xmax=1,
    ymin=0,
    ymax=1,
    axis equal image,
    hide axis,
    width=6.5cm
    }}
    
\newcommand*{\nodeFont}{\sansmath\sffamily\small}

%% file: abstract.tex
High-dimensional Nearest Neighbor (NN) search is central in multimedia search systems. Product Quantization (PQ) is a widespread NN search technique which has a high performance and a good scalability. PQ compresses high-dimensional vectors into compact codes thanks to a combination of quantizers. Large databases can therefore be stored entirely in RAM, enabling fast responses to NN queries. In almost all cases, PQ uses 8-bit quantizers as they offer low response times.

In this paper, we advocate the use of 16-bit quantizers. Compared to 8-bit quantizers, 16-bit quantizers boost accuracy but they increase response time by a factor of 3 to 10. We propose a novel approach that allows 16-bit quantizers to offer the same response time as 8-bit quantizers, while still providing a boost of accuracy. Our approach builds on two key ideas: (i) the construction of \emph{derived codebooks} that allow a fast and approximate distance evaluation, and (ii) a two-pass NN search procedure which builds a candidate set using the derived codebooks, and then refines it using 16-bit quantizers. On 1 billion SIFT vectors, with an inverted index, our approach offers a Recall@100 of 0.85 in 5.2 ms. By contrast, 16-bit quantizers alone offer a Recall@100 of 0.85 in 39 ms, and 8-bit quantizers a Recall@100 of 0.82 in 3.8 ms.

%% file: introduction.tex
%
%


%

%

Over the last decade, the amount of multimedia data handled by online services, such as video sharing web sites or social networks, has soared. This profusion of multimedia content calls for efficient multimedia search techniques, so as to allow users to find relevant content.
Efficient Nearest Neighbor (NN) search in high dimensionality is key in multimedia search. High-dimensional feature vectors, or descriptors, can be extracted from multimedia files, capturing their semantic content. Finding similar multimedia objects then consists in finding objects with a similar set of descriptors.

In low dimensionality, efficient solutions to the NN search problem, such as KD-trees, have been proposed. However, \emph{exact} NN search remains challenging in high-dimensional spaces. Therefore, current research work focuses on finding Approximate Nearest Neighbors (ANN) efficiently. Due to its good scalability and low response time, Product Quantization (PQ)~\cite{Jegou2011} is a popular approach \cite{Sun2013,Uchida2011,Yu2015} for ANN search in high-dimensional spaces. PQ compresses high-dimensional vectors into compact codes occupying only a few bytes. Therefore, large databases can be stored entirely in main memory. This enables responding to NN queries without performing slow I/O operations.

%
PQ compresses vectors into compact codes using a combination of quantizers. To answer ANN queries, PQ first pre-computes a set of lookup tables comprising the distance between the subvectors of the query vector and centroids of the quantizers. PQ then uses these lookup tables to compute the distance between the query vector and compact codes stored in the database. 
Previous work relies almost exclusively on 8-bit quantizers ($2^8$ centroids) which result in small lookup tables, that are both fast to pre-compute and fit the fastest CPU caches.


Novel quantization models inspired by Product Quantization (PQ) have been proposed to improve ANN search accuracy. These models, such as Additive Quantization (AQ)~\cite{Babenko2014AQ} or Tree Quantization (TQ)~\cite{Babenko2015TQ} offer a higher accuracy, but some of them lead to a much increased response time. An orthogonal approach to achieve a higher accuracy is to use 
16-bit quantizers instead of the common 8-bit quantizers. However, 16-bit quantizers are generally believed to be intractable because they result in a 3 to 10 times increase in response time compared to 8-bit quantizers



%
In this paper, we introduce a novel approach that makes 16-bit quantizers almost as fast as 8-bit quantizers while retaining the increase in accuracy. Our approach builds on two key ideas: (i) we build small \emph{derived codebooks} ($2^8$ centroids each) that approximate the codebooks of the 16-bit quantizers ($2^{16}$ centroids each) and (ii) we introduce a two-pass NN search procedure which builds a candidate set using the derived codebooks, and then refines it using the 16-bit quantizers. Moreover, our approach can be combined with inverted indexes, commonly used to manage large databases. More specifically, this paper makes the following contributions: 

\begin{itemize}
  \item We detail how we build derived codebooks that approximate the 16-bit quantizers. We describe how our two-pass NN search procedure exploits derived codebooks to achieve both a low response time and a high accuracy.

  \item We evaluate our approach in the context of Product Quantization (PQ) and Optimized Product Quantization (OPQ), with and without inverted indexes. We show that it offers close response time to 8-bit quantizers, while having the same accuracy as 16-bit quantizers.

  \item We discuss the strengths and limitations of our approach. We show that because it increases accuracy without significantly impacting response time, our approach compares favorably with the state of the art.
\end{itemize}





%% file: background.tex
In this section, we describe how Product Quantization (PQ) and Optimized Product Quantization (OPQ) encode vectors into compact codes. We then show how to perform ANN search in databases of encoded vectors.

\subsection{Vector Encoding}
\label{sec:vecenc}
Product Quantization (PQ)~\cite{Jegou2011} is based on the principle of vector quantization. A vector quantizer is a function $q$ which maps a vector $x \in \mathbb{R}^d$ to a vector $\mathcal{C}[i] \in \mathcal{C}$, where $\mathcal{C}$ is a predetermined set of $d$-dimensional vectors. The set of vectors $\mathcal{C}$ is named \emph{codebook}, and its elements are named \emph{centroids}. A quantizer which minimizes the quantization error maps a vector $x$ to its closest centroid:
\[
\operatorname{q}(x) = \argmin_{\mathcal{C}[i] \in \mathcal{C}}{||x-\mathcal{C}[i]||}.
\]
The index $i$ of the closest centroid $\mathcal{C}[i]$ of a vector $x \in \mathbb{R}^d$ can be used as a compact code representing the vector $x$: \(\operatorname{code}(x) = i\) such that \(\operatorname{q}(x) = \mathcal{C}[i]\).
The compact code $i$ only uses $b = \lceil \log_2(k) \rceil$ bits of memory, while a $d$-dimensional vector stored as an array of floats uses $d\cdot32$ bits of memory. The accuracy of ANN search primarily depends on the quantization error induced by the vector quantizer. Therefore, to minimize quantization error, it is necessary to use a quantizer with a large number of centroids, \eg $k = 2^{64}$ centroids. Yet, training such quantizers is not tractable.


Product Quantization addresses this issue by allowing to generate a quantizer with a large number of centroids from several low-complexity quantizers. To quantize a vector $x \in \mathbb{R}^d$, a product quantizer first splits $x$ into $m$ sub-vectors $x = (x^0,\dots,x^{m-1})$. Then, each sub-vector $x^j$ is  quantized using a distinct quantizer $q^j$. Each quantizer $q^j$ has a distinct codebook $\mathcal{C}^j$, of size $k$. A product quantizer maps a vector $x \in \mathbb{R}^d$ as follows:
\begin{align*}
\operatorname{pq}(x) &= \left ( \operatorname{q}^0(x^0),\dots,\operatorname{q}^{m-1}(x^{m-1}) \right )\\
&= (\mathcal{C}^0[i_0],\dots,\mathcal{C}^{m-1}[i_{m-1}])
\end{align*}
Thus, the codebook $\mathcal{C}$ of the product quantizer is the Cartesian product of the codebooks of the quantizers, $\mathcal{C} = \mathcal{C}^0 \times \dots \times \mathcal{C}^{m-1}$. The product quantizer has a codebook of size $k^m$, while only requiring training $m$ codebooks of size $k$. Product quantizers encode high-dimensional vectors into compact codes by concatenating the codes generated by the $m$ quantizers: \(\operatorname{pqcode}(x) = (i_0,\dots,i_{m-1}) \), such that \( \operatorname{q}(x) = (\mathcal{C}^0[i_0],\dots,\mathcal{C}^{m-1}[m-1])\).

Optimized Product Quantization (OPQ)~\cite{Ge2014} and Cartesian K-means (CKM)~\cite{Norouzi2013} are two similar quantizations schemes, both relying on the same principles as Product Quantization (PQ). An optimized product quantizer multiplies a vector $x \in \mathbb{R}^d$ by an orthonormal matrix $R \in \mathbb{R}^{d \times d}$ before quantizing it in the same way as a product quantizer: 
\[
\operatorname{opq}(x) = \operatorname{pq}(Rx), \text{ such that } R^TR=I
\]
The matrix $R$ allows an arbitrary rotation of the vector $x$, thus enabling a better distribution of the information between the quantizers. This, in turn, translates into a lower quantization error.

\subsection{Inverted Indexes}
\label{sec:invidx}
The simplest search strategy, known as \emph{exhaustive search}, consists in encoding database vectors as compact codes, and storing the codes as a contiguous array in RAM. To answer NN queries, the whole database is scanned~\cite{Jegou2011}.

The more elaborate \emph{non-exhaustive search} strategy builds on inverted indexes (or IVF) to avoid scanning the whole database at query time. An inverted index splits the vector space into K distinct cells. The compact codes of the vectors belonging to each cell are stored in a distinct inverted list. At query time, the closest cells to the query vector are determined and the corresponding inverted lists are scanned for nearest neighbors~\cite{Jegou2011,Babenko2015}. Because they allow scanning only a fraction of the database, inverted indexes offer a significant speed-up. In addition, most inverted indexes also increase accuracy thanks to residual encoding. This technique consists in encoding the residual $r(x)$ of a vector $x \in \mathbb{R}^d$ instead of encoding the vector $x$ itself. The residual $r(x)$ of a vector $x$ is defined as: $r(x) = x - c$, where $c$ is the center vector of the cell $x$ has been assigned to. Because they offer both a decrease in response time and an increase in accuracy, inverted indexes are widely used. They are especially useful for large databases, as exhaustive search is hardly tractable in this case.

\subsection{ANN Search}
\label{sec:annsearch}

\begin{algorithm}[t]
  \caption{ANN Search}\label{alg:ann}
  \begin{algorithmic}[1]
    \Function{nns}{$\{\mathcal{C}^j\}_{j=0}^{m}, \mathit{database}, y, r$}
    \State $\mathit{list} \gets$ \Call{index\_list}{$\mathit{database}, y$} \Comment{\emph{Index}}
    \State $\{D^j\}_{j=0}^{m} \gets$ \Call{compute\_tables}{$y, \{\mathcal{C}^j\}_{j=0}^{m}$} \Comment{\emph{Tables}}
    \label{alg:line:step1} 
    \State \Return{\Call{scan}{$\mathit{list}, \{D^j\}_{j=0}^{m}$}} 	\label{alg:line:step2} \Comment{\emph{Scan}}
    \EndFunction
    
    \Function{scan}{$\mathit{list}, \{D^j\}_{j=0}^{m}, r$}
    \State $\mathit{neighbors} \gets \operatorname{binheap}(r)$\Comment{binary heap of size $r$} \label{alg:line:binheap}
    \For{$i \gets 0$ to $|\mathit{list}|-1$}  \label{alg:line:iter}
    \State $c \gets \mathit{list}[i]$\Comment{$i$th compact code}
    \State $d \gets$ \Call{adc}{$c, \{D^j\}_{j=0}^{m}$} \label{alg:line:dist}
    \State $\mathit{neighbors}.\operatorname{add}((i, d))$  \label{alg:line:tupleadd}
    \EndFor
    \State \Return{$\mathit{neighbors}$}
    \EndFunction
    
    \Function{adc}{$c, \{D^j\}_{j=0}^{m-1}$} \label{alg:line:adc}
    \State $d \gets 0$
    \For{$j \gets 0$ to $m$}
    \State $d \gets d + D^j[c[j]]$ \label{alg:line:ops}
    \EndFor
    \Return{$d$}
    \EndFunction
  \end{algorithmic}
\end{algorithm}

ANN Search in a database of compact codes takes three steps: \emph{Index}, where the inverted index is used to retrieve the most appropriate inverted lists, \emph{Tables}, where a set of lookup tables are pre-computed and \emph{Scan}, which involves computing the distance between the query vector and the compact codes stored in the inverted lists. The \emph{Index} step is only necessary when non-exhaustive search is used and skipped in the case of exhaustive search.

\emph{Index Step.} In this step, the closest cell to the query vector $y$ is determined using the inverted index, and the corresponding inverted list are retrieved. The residual $r(y)$ of the query vector is also computed. To obtain a high recall, multiple cells are scanned for nearest neighbors. We denote as $\mathit{ma}$ the number of scanned inverted lists. Each step is repeated $ma$ times: $\mathit{ma}$ inverted lists are retrieved (\emph{Index} step), $\mathit{ma}$ sets of lookup tables are computed (\emph{Tables} step) and $\mathit{ma}$ inverted lists are scanned for nearest neighbors (\emph{Scan} step).

\emph{Tables Step.} In this step, a set of $m$ lookup tables $\{D^j\}_{j=0}^{m}$ are computed, one for each quantizer used by the product quantizer (Algorithm~\ref{alg:ann}, line \ref{alg:line:step1}). The $j$th lookup table is composed of the distances between the $j$th sub-vector of the query vector $y$ and all centroids of the $j$th quantizer:
\begin{equation}
  D^j = \left( \left \lVert {y}^j - \mathcal{C}^j[0] \right\rVert^2,\dots, \left \lVert {y}^j - \mathcal{C}^j[k-1] \right \rVert^2\right) \label{eqn:tables}
\end{equation}
We omitted the definition of the \textsc{compute\_tables} function in Algorithm~\ref{alg:ann}, but it corresponds to an implementation of Equation~\ref{eqn:tables}.

\emph{Scan Step.} In this step, the retrieved inverted list is scanned for nearest neighbors (Algorithm~\ref{alg:ann}, line~\ref{alg:line:step2}). The \textsc{scan} function iterates over all compact codes stored in the list (Algorithm~\ref{alg:ann}, line~\ref{alg:line:iter}) and computes the distance between each code and the query vector using the \textsc{adc} function (Algorithm~\ref{alg:ann}, line~\ref{alg:line:dist}). The \textsc{adc} function is an implementation of the Asymmetric Distance Computation (ADC)~\cite{Jegou2011} method which computes the distance between a code $c$ and the query vector $y$ as follows:
\begin{equation}
\operatorname{adc}(y,c) = \sum_{j=0}^{m-1} D^{j}[c[j]] \label{eqn:adc1}
\end{equation}
Equation \ref{eqn:adc1} is equivalent to:
\begin{equation}
\operatorname{adc}(y,c) = \sum_{j=0}^{m-1} \left \lVert {y}^j - \mathcal{C}^j[c[j]] \right \rVert^2 \label{eqn:adc2}
\end{equation}
%
To compute the distance between the query vector $y$ and a code $c$, ADC therefore sums the distances between the $m$ sub-vectors of $y$ and the $m$ centroids associated with the code $c$.

When the number of codes in the database is greater than the number of centroids of each quantizer, \ie $|\mathit{database}| > k$, using lookup tables avoids recomputing the same $\left \lVert {y}^j - \mathcal{C}^j[c[j]] \right \rVert^2$ terms multiple times.
Once the distance $d$ between the code $c$ and the query vector $y$ has been computed, the tuple $(i,d)$ is added to a binary heap of size $r$ (Algorithm~\ref{alg:ann}, line~\ref{alg:line:tupleadd}). This binary heap holds the $r$ tuples with the lowest distance $d$; other tuples are discarded. 


\subsection{Impact of the use of 16-bit quantizers}
\label{sec:16bitimpact}

\begin{table}
  \centering
  \caption{Response time and accuracy\label{tbl:impact}}
  \begin{tabularx}{\linewidth}{llXlll}
    \toprule
    $m \stimes b$ & Cache & R@100 & Index & Tables & Scan\\
    \midrule
    \multicolumn{6}{c}{SIFT1M, 64-bit codes, no index}\\
    \midrule
    $8 \stimes 8$ & L1 & 92.3\% & - & < 0.1 ms & \textbf{2.6 ms}\\
    $4 \stimes 16$ & L3 & 96.0\% & - & 0.65 ms & \textbf{6.4 ms}\\
    \midrule
    \multicolumn{6}{c}{SIFT1B, 64-bit codes, inverted index}\\
    \midrule
    $8 \stimes 8$ & L1 & 78.8\% & 0.67 ms & \textbf{0.14 ms} & \textbf{3.0 ms}\\
    $4 \stimes 16$ & L3 & 83.8\% & 0.67 ms & \textbf{24 ms} & \textbf{15 ms}\\
    \bottomrule
  \end{tabularx}
\end{table}

A product quantizer is fully characterized by two parameters: $m$, the number of quantizers and $k$ the number of centroids per quantizer. We denote $b = \lceil \log_2(k) \rceil$ the number of bits per quantizer, and we denote $m{\stimes}b$ product quantizer a product quantizer employing $m$ quantizers of $b$ bits each.

An $m{\stimes}b$ product quantizer encodes high-dimensional vectors into short codes occupying $m{\cdot}b$ bits of memory. The $m{\cdot}b$ product impacts the accuracy of ANN search, and the memory use of the database. The higher the $m{\cdot}b$ product, the higher the accuracy, but also the higher the memory use. In practice, 64-bit codes (\ie $m{\cdot}b = 64$) are commonly used, but there has been a recent interest in using 32-bit codes~\cite{Babenko2014AQ}.

For a fixed $m{\cdot}b$ product (\eg $m{\cdot}b = 64$ or $m{\cdot}b = 32$), there is the option of using either 8-bit quantizers ($b = 8$) or 16-bit quantizers ($b = 16$). Thus, a product quantizer generating 64-bit codes ($m{\cdot}b = 64$) can be built using either 8 8-bit quantizers ($8{\stimes}8$ product quantizer) or 4 16-bit quantizers ($4{\stimes}16$ product quantizer). Any other arbitrary value of $b$ than 8 or 16 would theoretically be possible. However, during a distance computation the $m$ $b$-bit integers composing a compact code are accessed individually. For efficiency, it is important that these integers have a size that can be natively addressed by CPUs, \ie $b \in \{8,16,32,64\}$. In addition, 32-bit or 64-bit vector quantizers are not tractable, leaving $b = 8$ and $b = 16$ as practical values.

We measure the impact of the use of 16-bit quantizers on accuracy (using the Recall@100 measure, denoted R@100), and on the time spent in the Index, Tables and Scan steps (Table~\ref{tbl:impact}). We use a small dataset of 1 million SIFT vectors (SIFT1M), for which we do not use an inverted index, and a large dataset of 1 billion SIFT vectors (SIFT1B), for which we use an inverted index ($K=65536$ cells, $\mathit{ma}=64$). In both cases, 16-bit quantizers significantly boost accuracy, but they also cause an increase in the time spent in the Tables and Scan step. The use of 16-bit quantizers in place of 8-bit quantizers produces much larger $D^j$ lookup tables ($2^{16}$ floats each versus $2^8$ floats each), which has two main consequences: (i) the lookup tables are more costly to pre-compute ($2^{16}$ distance computations versus $2^8$), which increases Tables time, and (ii) the lookup tables have to be stored in larger but slower cache levels, which increases Scan time. For 8-bit quantizers, lookup tables fit in the fastest CPU cache, the L1 cache (4-5 cycles latency). For 16-bit quantizers, lookup tables have to be stored in the larger but slower L3 cache ($\approx50$ cycles latency).

%% file: joint_quantizers.tex
\begin{figure*}[t]%
  \subfloat[][Temp. codebook $C_j^t$]{%
    \input{fig/quantizer/quantizer0}
    \label{fig:quant0}
  }\hfil%
  \subfloat[][Partition $\overline{P}$ of $C_j^t$]{%
    \input{fig/quantizer/quantizer1}
    \label{fig:quant1}
  }\hfil%
  \subfloat[][Final quantizer ($C_j$)]{%
    \input{fig/quantizer/quantizer2}
    \label{fig:quant2}
  }\hfil%
  \subfloat[][Derived codebook ($\overline{C_j}$)]{%
    \input{fig/quantizer/quantizer3}
    \label{fig:quant3}
  }\hfil%
  \caption{Derived codebook training process}
  \label{fig:vecgroup}%
\end{figure*}
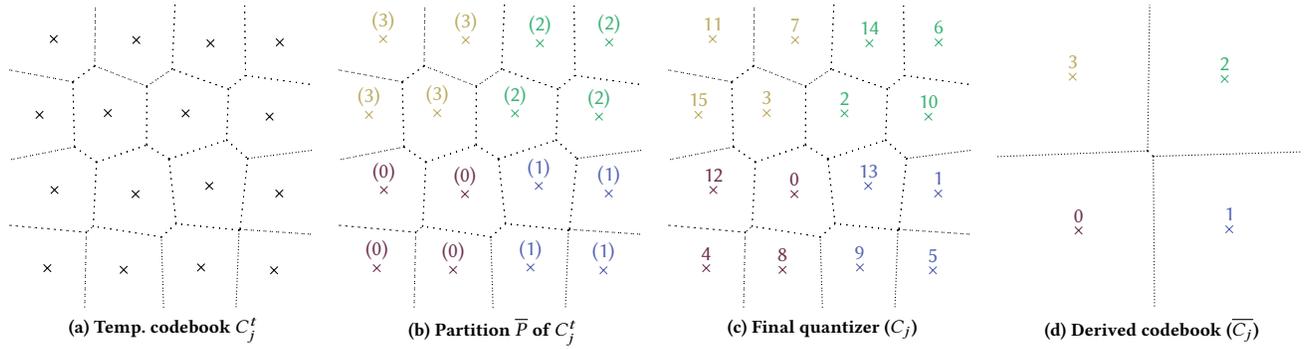

\subsection{Overview}

%
%

Thanks to the use of derived codebooks, our approach allows both strongly limiting the increase in Tables time and the increase in Scan time caused by the use of 16-bit quantizers. Our approach therefore performs well with and without inverted indexes. The gist of our approach is to build small \emph{derived codebooks} ($2^{8}$ centroids each) that approximate the codebooks of the 16-bit quantizers, and to exploit these derived codebooks to speedup ANN search. We use the derived codebooks to compute \emph{small lookup tables} that fit the fast L1 cache, and that can be used for approximate distance evaluation. We build a candidate set of $r2$ vectors by computing approximate distances. This candidate set is then refined by performing a precise distance evaluation, which relies on the codebooks of the 16-bit quantizers. 

Therefore, with the conventional approach, ANN search takes two steps: \emph{Tables} and \emph{Scan} (with an optional \emph{Index} step, Section \ref{sec:annsearch}). With our approach, ANN search takes three steps: \emph{Tables}, during which we pre-compute small lookup tables, \emph{Scan} during which we build a candidate set using small lookup tables and \emph{Refine}, during which we refine the candidate set using the 16-bit quantizers. An optional \emph{Index} step can also be used with our approach.
In the remainder of this paper, we denote \pq{m}{\overline{b},b} a product quantizer with $m$ quantizers of $b$ bits each, associated with $m$ derived codebooks of $2^{\overline{b}}$ 
centroids each (respectively O\pq{m}{\overline{b},b} an optimized product quantizer with similar properties). We focus on (O)\pq{m}{8,16}, \ie (optimized) product quantizers with 16-bit quantizers and derived codebooks of $2^{8}$ centroids.

\subsection{Building Derived Codebooks}

\begin{algorithm}[t]
  \caption{Building derived codebooks\label{alg:jointbuild}}
  \begin{algorithmic}[1]
    \Function{build\_quantizers}{$V_t,  \overline{k} , k$}
    \State $\mathcal{C}_j^t, P\gets$ \Call{kmeans}{$V_t, k$}
    \Comment{Step 1}
    \State $\overline{\mathcal{C}_j}, \overline{P} \gets$ \Call{kmeans\_same\_size}{$\mathcal{C}_j^t, \overline{k}$}
    \Comment{Step 2}
    \State $\mathcal{C}_j \gets$ \Call{build\_final\_codebook}{$\overline{P}$, $\overline{b}$}
    \Comment{Step 3}
    \State \Return{$\mathcal{C}_j, \mathcal{C}_j'$}
    \EndFunction
    \Function{build\_final\_codebook}{$\overline{P}$, $\overline{k}$}
    \State $\overline{b} = \log_2(k)$
    \For{$l \gets 0$ to $\overline{k}$}
    \State $\overline{G} \gets \overline{P[i]}$
    \For{$\overline{i} \gets 0$ to $|\overline{G}|-1$}
    \State $\mathcal{C}_j[\overline{i} \ll \overline{b} \mathrel{|} l] = \overline{G}[\overline{i}]$ \Comment{\small{$|$ is binary or}}
    \Statex \Comment{\small{$\ll$ is bitwise left shift}} \label{alg:line:order}
    \normalsize
    \EndFor
    \EndFor
    \State \Return{$\mathcal{C}_j$}
    \EndFunction
  \end{algorithmic}
\end{algorithm}

A product quantizer uses $m$ quantizers, each having a different codebook $\mathcal{C}_j, j \in \{0,\dots,m-1\}$. We build the derived codebook $\overline{\mathcal{C}_j}, j \in \{0,\dots,m-1\}$ from the codebook of the corresponding quantizer. Thus, $\mathcal{C}_0$ is derived from  $\overline{\mathcal{C}_0}$ \etc

%
The training process of the codebook of the $j$th quantizer, with $k=2^{b}$ centroids, and of the $j$th derived codebook, with $\overline{k}=2^{\overline{b}}$ centroids is described in Algorithm~\ref{alg:jointbuild}. The \textsc{kmeans} function is a standard implementation of the k-means algorithm. It takes a training set $V_t$ and a parameter $k$, the desired number of clusters. It returns a codebook $\mathcal{C}$ and a partition $P$ of the training set into $k$ clusters. We denote $P[i]$ the $i$th cluster of $P$. The \textsc{kmeans\_same\_size} function is a k-means variant \cite{SSZKmeans} which produces clusters $G$ of identical sizes, \ie $\forall G_l \in P, |G_l| = |G_0|$ . For the sake of simplicity, Figure~\ref{fig:vecgroup} illustrates the training process for $k=16$ and $\overline{k} = 4$, although we use $k=2^{16}$ and $\overline{k}=2^8$ in practice. The training process takes three steps, described in the three following paragraphs.

\emph{Step 1.} Train a temporary codebook $\mathcal{C}_j^t$ using the \textsc{kmeans} function. Figure~\ref{fig:quant0} shows the result of this step. Each point represents a centroid of $\mathcal{C}_j^t$. Vectors of the training set $V_t$ are not shown. Implicitly, centroids of $\mathcal{C}_j^t$ have an index associated with them, which is their position in the list $\mathcal{C}_j^t$, \ie the index of $\mathcal{C}_j^t[i]$ is i. As the indexes of centroids are not used in the remainder of the training process, they are not shown. 

\emph{Step 2.} Partition $\mathcal{C}_j^t$ into $\overline{k}$ clusters using \textsc{kmeans\_}\textsc{same\_} \textsc{size}. To do so, $\mathcal{C}_j^t$ is used as the training set for \textsc{kmeans\_same} \textsc{\_size}. Figure~\ref{fig:quant1} shows the partition $\overline{P}=(\overline{G_l}), l \in \{0,\dots,\overline{k}-1\}$ of $\mathcal{C}_j^t$. The number in parentheses above each centroid in the index of the cluster it has been assigned to \ie the number $l$ such that the centroid belongs to $\overline{G_l}$. The codebook $\overline{\mathcal{C}_j}$ returned by \textsc{kmeans\_same\_size} is the derived codebook, shown on Figure~\ref{fig:quant3}. Each centroid $\mathcal{C}_j[l]$ of the derived codebook is the centroid of the cluster $\overline{G_l}$. 

\emph{Step 3.} Build the final codebook $\mathcal{C}_j$ by reordering the centroids of the temporary codebook $\mathcal{C}_j^t$. \emph{This reordering, or re-assignment of centroid indexes, is the key that allows the derived codebooks $\overline{\mathcal{C}_j}$ to be used as an approximate version of the $\mathcal{C}_j$ codebooks}. The order of centroids in $\mathcal{C}_j$ must be such that the lowest $\overline{b}$ bits of the index assigned to each centroid of $\mathcal{C}_j$ matches the cluster $\overline{G_l}$ it has been assigned to in step 2. If we denote $\operatorname{low_{\overline{b}}}(i)$, the lower $b$ bits of the index $i$, the order of centroids must obey the property:
\begin{gather*}
  \forall {i \in \{0..k-1\}}, \forall l \in \{0..\overline{k}-1\},\\
  \operatorname{low_{\overline{b}}}(i) = l \Leftrightarrow \mathcal{C}_j[i] \in G_l \tag{P1}
\end{gather*}
The \textsc{build\_final\_codebook} function produces an assignment of centroid indexes which obeys property P1 (Algorithm~\ref{alg:jointbuild}, line~\ref{alg:line:order}). Figure~\ref{fig:quant2} shows the final assignment of centroid indexes. In this example, $k=16$ and $\overline{k}=4$ ($b=4$ and $\overline{b}=2$). The centroids belonging to cluster 1 (\texttt{01} in binary) have been assigned the indexes 9 (\texttt{1001}), 13 (\texttt{1101}), 1 (\texttt{0001}), and 5 (\texttt{0101}). The lowest $\overline{b}=2$ bits of 9,13,1 and 5 are \texttt{01}, which matches the partition number (1, or \texttt{01}). This property similarly holds for all partitions and all centroids.

This joint training process allows the derived codebook $\overline{\mathcal{C}_j}$ to be used as an approximate version of $\mathcal{C}_j$ during the NN search process. To encode a vector $x \in \mathbb{R}^d$ into a compact code, the codebooks $\mathcal{C}_j$ are used. The code $c$ resulting from the encoding of vector $x$ is such that for all $j \in \{0,\dots,m-1\}$, $\mathcal{C}_j[c[j]]$ is the closest centroid of $x^j$ in $\mathcal{C}_j$. Our training process ensures that the centroid $\overline{\mathcal{C}_j}[\operatorname{low_{\overline{b}}}(c[j])]$ is close to $x^j$. In other words, the centroid index assigned by the quantizer $\mathcal{C}_j$ remains meaningful in the derived codebook $\overline{\mathcal{C}_j}$.




\subsection{ANN Search with Derived Codebooks}
\label{sec:budsearch}
ANN Search with derived codebooks takes four steps: \emph{Index}, where the appropriate inverted lists are retrieved from the inverted index, \emph{Tables}, where small lookup tables are computed from the derived codebooks, \emph{Scan}, where a candidate set if $r2$ vectors is built using the small lookup tables and \emph{Refine}, where the candidate set is refined using the 16-bit quantizers. The \emph{Index} step is identical to the \emph{Index} step of the conventional ANN Search procedure (Section~\ref{sec:annsearch}), therefore we do not detail it here. All other steps (Tables, Scan and Refine) are different, and described below: 
%
%

\begin{figure}[t]
  \centering
  \input{fig/buckets.tex}
  \caption{Capped buckets data structure}
\end{figure}
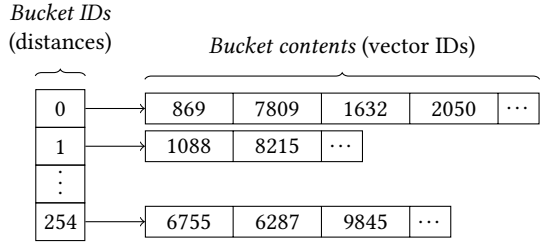

\pgfplotsset{small fonts/.style={
    title style = {font=\small, yshift=-0.5em},
    label style = {font=\small},
    tick label style = {font=\small},
    legend style={font=\small}
}}

\pgfplotsset{sfit1mr2 plot/.style={
    small fonts,
    xtick={2000,5000,10000,20000,50000},
    xticklabels={2K,5K,10K,20K,50K},
}}

\pgfplotsset{tiny10mr2 plot/.style={
    small fonts,
    xtick={20000,50000,100000,200000,500000},
    xticklabels={20K,50K,100K,200K,500K},
}}


\pgfplotsset{r plot/.style={
    small fonts,
    xtick={10,20,50,100,200,500,1000},
    xticklabels={10,20,50,100,200,500,1K},
    xlabel=$r$,
    width=0.48*\linewidth,
    height=0.32*\linewidth,
    legend cell align=left
}}

\pgfplotsset{r plot with dual axis/.style={
  small fonts,
  xtick={10,20,50,100,200,500,1000},
  xticklabels={10,20,50,100,200,500,1K},
  tick label style = {font=\small, align=left, inner ysep=0.7em, 
    fill=none, inner xsep=0},
  extra x ticks={10},
  extra x tick style={tick label style={
      anchor=north east, align=right, inner xsep=1em, inner ysep=0.7em}},
  extra x tick labels={$r\rightarrow$\\[0.2em]$r2\rightarrow$},
  width=0.35*\linewidth,
  height=0.27*\linewidth,
  legend cell align=left
}}


%
\emph{Tables Step.} In this step, a set of $m$ \emph{small lookup tables}, $\{\overline{D_j}\}_{j=0}^{m}$ are computed from the derived codebooks $\{\overline{\mathcal{C}_j}\}_{j=0}^{m}$. Small lookup tables are computed using the same \textsc{compute\_tables} function as the one used in the conventional ANN search procedure (Section \ref{sec:annsearch}, Algorithm \ref{alg:ann}). The short lookup table $\overline{D_j}$ consists of the distances between the sub-vectors $y^j$ and all centroids of the codebook $\overline{\mathcal{C}_j}$.
Unlike in the conventional ANN search procedure, we then quantize the floating-point distances in small lookup tables $\{\overline{D}_j\}_{j=0}^{m-1}$ to 8-bit integers, in order to build the quantized lookup tables $\{Q_j\}_{j=0}^{m-1}$. This additional quantization stage is necessary because our scan procedure (\emph{Scan} Step) uses a data structure  optimized for fast insertion, capped buckets, which requires distances to be quantized to 8-bit integers. We follow a quantization procedure similar to the one used in~\cite{Andre2015}. We quantize floating-point distances uniformly into 255 (0-254) bins between a $\mathit{qmin}$ and $\mathit{qmax}$ bound. All distances above $\mathit{qmax}$ are quantized to 255. We use the minimum distance across all lookup tables as the $\mathit{qmin}$ bound. To determine $\mathit{qmax}$, we compute the distance between the query vector and the $r2$ first vectors of the database. The greatest distance is used as the $\mathit{qmax}$ bound. Once $\mathit{qmin}$ and $\mathit{qmax}$ have been set, quantized lookup tables are computed as follows:
\begin{gather*}
\forall j \in \{0,\dots,m-1\}, \forall i \in \{0, \dots, \overline{k}-1\}, \\
Q_j[i] = \left \lfloor \frac{\overline{D_j}[i] - \mathit{qmin}}{\mathit{qmax} - \mathit{qmin}}  \cdot 255 \right \rfloor
\end{gather*}

\begin{algorithm}[t]
  \caption{ANN Search with derived codebooks}\label{alg:annbud}
  \begin{algorithmic}[1]
    \Function{nns\_derived}{$\{\mathcal{C}^j\}_{j=0}^{m}, \{\overline{\mathcal{C}^j}\}_{j=0}^{m} ,\mathit{db}, y, r, r2$}
    \State $\mathit{list} \gets$ \Call{index\_list}{$\mathit{db}, y$} \Comment{\emph{Index}}
    \State $\{\overline{D^j}\}_{j=0}^{m} \gets$ \Call{compute\_tables}{$y, \{\overline{\mathcal{C}^j}\}_{j=0}^{m}$} \label{alg:line:bstep1}\Comment{\emph{Tables}}
    \State $\{Q_j\}_{j=0}^m \gets$ \Call{quantize}{$\{\overline{D^j}\}_{j=0}^{m}, list, r2$} \label{alg:line:bstep2}
    \State $\mathit{cand} \gets $ \Call{scan}{$list, \{Q_j\}_{j=0}^m, r2$}  \label{alg:line:bstep3}\Comment{\emph{Scan}}
    \State \Return{\Call{refine}{$\mathit{list}, \mathit{cand},\{\mathcal{C}^j\}_{j=0}^{m},y,r,r2$}}\Comment{\emph{Refine}}
    \label{alg:line:bstep4}
    \EndFunction
    
    \Function{scan}{$\mathit{list}, \{Q_j\}_{j=0}^m, r2$} \label{alg:line:bscan}
    \State $\mathit{cand} \gets \operatorname{capped\_buckets}(r2)$ \label{alg:line:bucketinst}
    \For{$i \gets 0$ to $|\mathit{list}|-1$}
    \State $c \gets \mathit{list}[i]$\Comment{$i$th compact code}
    \State $d \gets$ \Call{adc\_low\_bits}{$c, \{Q^j\}_{j=0}^{m}$}
    \State $\mathit{cand}.\operatorname{put}(d, i)$ \label{alg:line:bucketput}
    \EndFor
    \State \Return{$\mathit{cand}$}
    \EndFunction
    
    \Function{adc\_low\_bits}{$c, \{Q_j\}_{j=0}^m$}
    \State $d \gets 0$
    \For{$j \gets 0$ to $m-1$}
    \State $d \gets d + Q_j[\operatorname{low_{\overline{b}}}(c[j])]$ \label{alg:line:qaccess}
    \EndFor
    \State \Return $d$
    \EndFunction
    
    \Function{refine}{$\mathit{list},\mathit{cand},\{\mathcal{C}^j\}_{j=0}^{m},y,r,r2$}
    \State $i_{bucket} \gets 0$
    \State $\mathit{count} \gets 0$
    \State $\mathit{neighbors} \gets \operatorname{binheap}(r)$ \label{alg:line:finalset}
    \State $\{D_j\}_{j=0}^{m-1} \gets \{\{-1\}\}$ \label{alg:line:tblfill}
    \While{$count < r2$}
    \State $\mathit{bucket} \gets \mathit{cand}.\operatorname{get\_bucket}(i_{bucket})$ \label{alg:line:bucketiter}
    \ForAll{$i \in \mathit{bucket}$} \label{alg:line:bucketproc}
    \State $d \gets$ \Call{adc\_refine}{$\mathit{list}[i], \{D^j\}_{j=0}^{m-1}, \{\mathcal{C}^j\}_{j=0}^{m}$}
    \State $\mathit{neighbors}.\operatorname{add}((i, d))$
    \EndFor
    \State $\mathit{count} \gets \mathit{count} + \mathit{bucket}$.size
    \State $i_{bucket} \gets i_{bucket} + 1$ \label{alg:line:bucketiter2}
    \EndWhile
    \EndFunction
    
    \Function{adc\_refine}{$c, \{D^j\}_{j=0}^{m-1}, \{\mathcal{C}_j\}_{j=0}^{m-1}$}
    \State $d \gets 0$
    \For{$j \gets 0$ to $m-1$}
    \If{$D^j[c[j]] = -1$} \label{alg:line:notinit}
    \State $D^j[c[j]] \gets \left \lVert y^j - \mathcal{C}^j[c[j]] \right \rVert^2$ \label{alg:line:calctbl}
    \EndIf
    \State $d \gets d + D^j[c[j]]$
    \EndFor
    \Return{$d$}
    \EndFunction
  \end{algorithmic}
\end{algorithm}

\emph{Scan Step.} In this step, the full database is scanned to build a candidate set of $r2$ vectors. This candidate set is built by performing approximate distance evaluations which rely on the quantized lookup tables (Algorithm~\ref{alg:annbud}, line~\ref{alg:line:bscan}). Our scan procedure is similar to the scan procedure of the conventional ANN search algorithm (Algorithm~\ref{alg:ann}), apart from two differences. First, the \textsc{adc\_low\_bits} function is used to compute distances in place of the \textsc{adc} function. The \textsc{adc\_low\_bits} function masks the lowest $\overline{b}$ bits of centroids indexes $c[j]$ to perform lookups in the small lookup tables (Algorithm~\ref{alg:annbud}, line~\ref{alg:line:qaccess}), instead of using the full centroids indexes to access the full lookup tables. Second, candidates are stored in a data structure optimized for fast insertion, capped buckets (Algorithm~\ref{alg:annbud}, line~\ref{alg:line:bucketinst}), instead of a binary heap. Capped buckets consists of an array of buckets, one for each possible distance (0-254). Each bucket is a list of vector IDs. The put operation (Algorithm~\ref{alg:annbud}, line~\ref{alg:line:bucketput}) involves retrieving the bucket $d$, and appending the vector ID $i$ to the list.
Because adding a vector to a capped buckets data structure requires much less operations than adding a vector to a binary heap, it is much faster ($\operatorname{O}(1)$ operations versus $\operatorname{O}(\log{n})$ operations). It however requires distances to be quantized to 8-bit integers, which are used as bucket IDs. This not an issue, as quantizing distances to 8-bit integers has been shown not to impact recall significantly~\cite{Andre2015}. On the contrary, a fast insertion is highly beneficial because the candidate set is relatively large (typical $r2$=10K-200K) in comparison with the final result set (typical $r$=10-100). Therefore, many insertions are performed in the candidate set.
To avoid the capped buckets data structure to grow indefinitely, we maintain an upper bound on distances; vectors having distances higher than the upper bound are discarded. The distance (\ie bucket ID) of the $r2$-th farthest vector in the capped buckets is used as upper bound. 

%
\emph{Refine Step.} In this step, $r2$ vectors are extracted from the capped buckets, and a precise distance evaluation is performed for these vectors. These precise distance evaluations use the full quantizers $\{\mathcal{C}_j\}_{j=0}^{m-1}$. Precise distances are used to build the result set, denoted $\mathit{neighbors}$ (Algorithm~\ref{alg:annbud}, line~\ref{alg:line:finalset}). We iterate over capped buckets by increasing bucket ID $i_{bucket}$ (Algorithm~\ref{alg:annbud}, line~\ref{alg:line:bucketiter} and line~\ref{alg:line:bucketiter2}). We process bucket 0, bucket 1, \etc until $r2$ vectors have been processed. When processing a bucket, we iterate over all vectors IDs stored in this bucket (Algorithm~\ref{alg:annbud}, line~\ref{alg:line:bucketproc}), and compute precise distances using the \textsc{adc\_refine} function. This function is similar to the \textsc{adc} function used in the conventional search process (Algorithm~\ref{alg:ann}). However, here, we do not pre-compute the full lookup tables $\{D_j\}_{j=0}^{m-1}$ but rely on a dynamic programming technique. We fill all tables with the value -1 (Algorithm~\ref{alg:annbud}, line~\ref{alg:line:tblfill}), and compute table elements on demand. Whenever, the value -1 is encountered during a distance computation (Algorithm~\ref{alg:annbud}, line~\ref{alg:line:notinit}), it means that this table element has not yet be computed. The appropriate centroid to sub-vector distance is therefore computed and stored in the lookup tables (Algorithm~\ref{alg:annbud}, line~\ref{alg:line:calctbl}). This strategy is beneficial because it avoids computing the full $\{D_j\}_{j=0}^{m-1}$ tables, which is costly (Table~\ref{tbl:impact}). In the case of \pq{m}{8,16}, a full lookup table $D_j$ comprises a large number of elements $k=2^{16}=65536$, but only a small number are accessed (usually 5\%-20\%).

%% file: fig/quantizer/quantizer0.tex
\begin{tikzpicture}
\begin{axis}[voronoiaxis]
\addplot [
    only marks, mark=x]
    table {fig/quantizer/quantizer_centroids.dat};
\addplot [no markers, update limits=false, dotted] table {fig/quantizer/quantizer_voronoi.dat};

\end{axis}
\end{tikzpicture}

%% file: fig/quantizer/quantizer1.tex
\begin{tikzpicture}
\begin{axis}[voronoiaxis]
\addplot [
    only marks, mark=x, 
    scatter, scatter src=explicit,
    colormap name=rainbow16,
    visualization depends on={\thisrow{group} \as \group},
    nodes near coords*={(\pgfmathprintnumber[int trunc]{\pgfplotspointmeta})},
    every node near coord/.append style={font=\nodeFont, adjust near node color},
    point meta=\thisrow{group},
    ]
    table {fig/quantizer/quantizer_centroids.dat};
\addplot [no markers, update limits=false, dotted] table {fig/quantizer/quantizer_voronoi.dat};

\end{axis}
\end{tikzpicture}

%% file: fig/quantizer/quantizer2.tex
\begin{tikzpicture}
\begin{axis}[voronoiaxis]
\addplot [
    only marks, mark=x, 
    scatter, scatter src=explicit,
    colormap name=rainbow16,
    visualization depends on={\thisrow{label} \as \pointLabel},
    nodes near coords*={$\pgfmathprintnumber[int trunc]{\pointLabel}$},
    every node near coord/.append style={font=\nodeFont, adjust near node color},
    point meta=\thisrow{group},
    ]
    table {fig/quantizer/quantizer_centroids.dat};
\addplot [no markers, update limits=false, dotted] table {fig/quantizer/quantizer_voronoi.dat};

\end{axis}
\end{tikzpicture}

%% file: fig/quantizer/quantizer3.tex
\begin{tikzpicture}
\begin{axis}[voronoiaxis]
\addplot [
    only marks, mark=x, 
    scatter, scatter src=explicit,
    colormap name=rainbow16,
    nodes near coords*={\pgfmathprintnumber[int trunc]{\pgfplotspointmeta}},
    every node near coord/.append style={font=\nodeFont, adjust near node color},
    point meta=\thisrow{label},
    ]
    table {fig/quantizer/quantizer_cheap_centroids.dat};
\addplot [no markers, update limits=false, dotted] table {fig/quantizer/quantizer_cheap_voronoi.dat};

\end{axis}
\end{tikzpicture}

%% file: fig/buckets.tex
\newcommand{\idHeight}{2eX}

\newlength\cellWidth
\setlength{\cellWidth}{\widthof{254}}

\tikzset{
  array cell/.style = {
    rectangle,
    draw,
    text width=\cellWidth,
    minimum height=3.7ex,
    align=center,
  },
  array/.style = {
    matrix of math nodes,
    row sep = -\pgflinewidth,
    column sep = -\pgflinewidth,
    nodes={
      array cell
    },
  }
}


\tikzstyle{bucket} = [
  rectangle split,
  rectangle split horizontal,
  draw
]

\newcommand{\slotWidth}{3em}

\tikzstyle{slot} = [
  align=center,
  text width=\slotWidth
]

\newcommand{\bucketSep}{2.5em}

\makeatletter
\DeclareRobustCommand{\rvdots}{%
  \vbox{
    \baselineskip4\p@\lineskiplimit\z@
    \kern-\p@
    \hbox{.}\hbox{.}\hbox{.}
  }}
\makeatother

\begin{tikzpicture}

\node [array] (distances)
{%
  0\\
  1\\
  \rvdots\\
  254\\
};

\node [bucket,
rectangle split parts=5,
right=\bucketSep of distances-1-1.east,
anchor=west] (bucket0) {%
  \nodepart[slot]{one}869
  \nodepart[slot]{two}7809
  \nodepart[slot]{three}1632
  \nodepart[slot]{four}2050
  \nodepart{five}\dots
};

\draw[->] (distances-1-1.east) -- (bucket0.west);

\node [bucket,
  rectangle split parts=3,
  right=\bucketSep of distances-2-1.east,
  anchor=west] (bucket1) {%
  \nodepart[slot]{one}1088
  \nodepart[slot]{two}8215
  \nodepart{three}\dots
};

\draw[->] (distances-2-1.east) -- (bucket1.west);

\node [bucket,
rectangle split parts=4,
right=\bucketSep of distances-4-1.east,
anchor=west] (bucket254) {%
  \nodepart[slot]{one}6755
  \nodepart[slot]{two}6287
  \nodepart[slot]{three}9845
  \nodepart{four}\dots
};

\draw[->] (distances-4-1.east) -- (bucket254.west);

\draw [decoration={brace,raise=5pt},
  decorate] (distances-1-1.north west) --  node[above=10pt,align=center] {\emph{Bucket IDs}\\(distances)} (distances-1-1.north east);
  
\draw [decoration={brace,raise=5pt},
decorate] (bucket0.north west) --  node[above=10pt] {\emph{Bucket contents} (vector IDs)} (bucket0.north east);

\end{tikzpicture}

%% file: evaluation.tex
\label{sec:eval}

\subsection{Experimental Setup}
All ANN search methods evaluated in this section are implemented in C++. We use the g++ compiler version 6.3, with the options \texttt{-O3} \texttt{-ffast-math} \texttt{-m64 -march=native}. For linear algebra primitives, we use OpenBLAS version 0.2.19, of which we compiled an optimized version on our system. We trained the full codebooks $\mathcal{C}_j$ of product quantizers and optimized product quantizers using the implementation\footnote{\url{https://github.com/arbabenko/Quantizations}} of the authors of~\cite{Babenko2014AQ, Babenko2015TQ}. We perform experiments on three datasets: 
\begin{itemize}
  \item SIFT1M\footnote{\label{foot:texmex}\url{http://corpus-texmex.irisa.fr/}}, a dataset of 1 million SIFT vectors (128 dimensions) with a training set of 100 thousand vectors
  \item TINY10M, a dataset of 10 million GIST vectors (384 dimensions) with a training set of 10 million vectors, extracted from the TINY dataset\footnote{\url{http://horatio.cs.nyu.edu/mit/tiny/data/index.html}}
  \item SIFT1B\cref{foot:texmex}, a large-scale dataset of 1 billion SIFT vectors. We use a training set of 3 million vectors.
\end{itemize}
To characterize the behavior of our approach, we first evaluate in the context of exhaustive search (\ie without inverted index) on the small SIFT1M and TINY10M datasets. We then show that our approach also performs well in the context of non-exhaustive search (\ie with an inverted index) on the SIFT1B dataset. All experiments were performed on a workstation equipped with an Intel Xeon E5-1650v3 CPU and 16GiB of RAM (DDR4 2133Mhz).




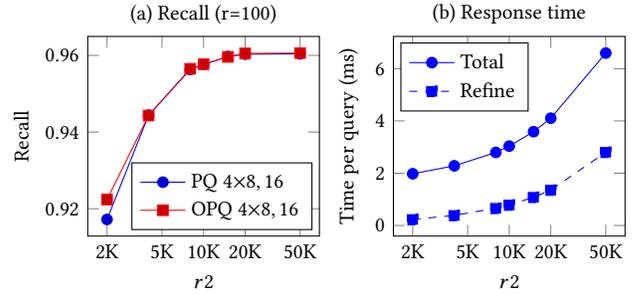
\begin{figure}[t]
  \centering
  \input{graphs/SIFT1M_r2.tex}
  \caption{Impact of $r2$ (SIFT1M, 64-bit codes)\label{fig:sift1m_r2}}
\end{figure}

\begin{figure*}[t]
  \centering
  \input{graphs/time_recall_graphs.tex}
  \caption{Recall and response time (64-bit codes)\label{fig:64ball}}
\end{figure*}
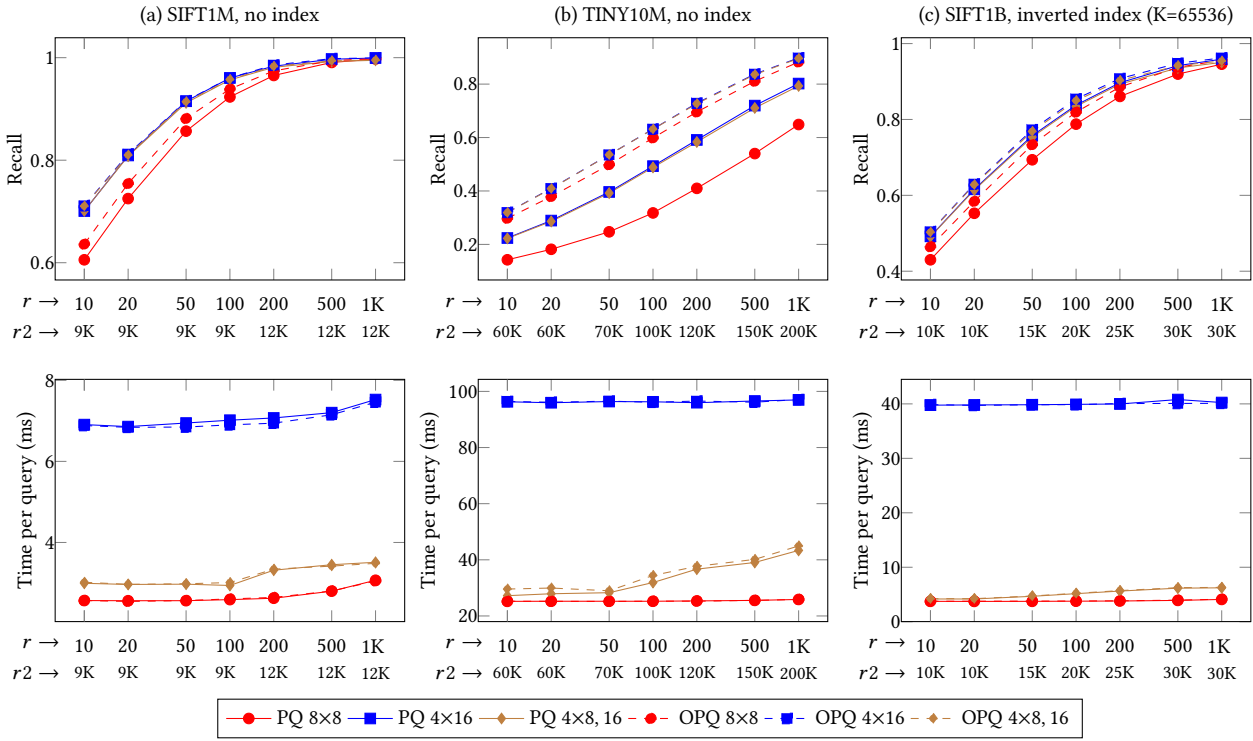

\subsection{Impact of r2}

The recall of the conventional ANN search method depends only on the $r$ parameter, the size of the result set. Our approach depends on an additional parameter, $r2$ the size of the candidate set. To assess the impact of $r2$, we set $r=100$ and measure the impact of $r2$ on recall and response time (Figure~\ref{fig:sift1m_r2}). Recall increases with $r2$ until it reaches a plateau (Figure~\ref{fig:sift1m_r2}a). Response time strongly increases with $r2$ (Figure~\ref{fig:sift1m_r2}b). This is mainly because the Refine step of our procedure becomes much more costly as $r2$ increases. A high $r2$ means that many costly subvector-to-centroid distance computations will be performed, and that many accesses to slow cache levels will be performed. In addition, the Scan step also becomes slightly more costly as $r2$ increases, because more insertions in the capped buckets structure are performed. Therefore, it is important that $r2$ is high enough so that our approach (denoted PQ $4{\stimes}8,16$ for product quantization and OPQ $4{\stimes}8,16$ for optimized product quantization) achieves the same recall as 16-bit quantizers (denoted PQ $4{\stimes}16$, OPQ $4{\stimes}16$), but not too high so as not to increase response time too much. For each value of $r$, we select the lowest $r2$ such that the recall of (O)PQ$4{\stimes}8,16$ is within 1\% of the recall of (O)PQ$4{\stimes}16$. We determine the appropriate $r2$ value using a subset of the query set. We report both values, $r$ and corresponding $r2$, on the x-axis of our graphs (Figure~\ref{fig:64ball}).

\subsection{Exhaustive search, 64-bit codes}
On the SIFT1M dataset, our approach achieves the same recall as 16-bit quantizers, for a response time comparable to the one of 8-bit quantizers (Figure~\ref{fig:64ball}a). While 16-bit quantizers incur a 2.7 times increase in response time over 8-bit quantizers, our approach achieves the same recall for only a 1.1 times increase in response time. The response time of our approach increases for $r > 100$, because we increased $r2$ from 9000 to 12000 to maintain the same recall as 16-bit quantizers.

For the TINY10M dataset (GIST descriptors), the benefit of using 16-bit quantizers instead of 8-bit quantizers is lower than it is for SIFT descriptors, especially in the case of OPQ (Figure~\ref{fig:64ball}b). Yet, our approach also achieves the same recall as 16-bit quantizers on this dataset. The response time of our approach again increases with $r2$. Nonetheless, our approach has a 1.07 times higher response time than 8-bit quantizer in the best case, and a 1.68 higher response time in the worst case. On the contrary, 16-bit quantizers incur a 2.8 times increase in response time in all cases (Figure~\ref{fig:64ball}b).

\subsection{Non-exhaustive search, 64-bit codes}


So far, we have evaluated our approach in context of exhaustive search, with small datasets. We now evaluate our approach on a large dataset of 1 billion vectors and with an inverted index. To demonstrate the applicability of our approach in this context, we combine it with a simple inverted index (IVFADC, \cite{Jegou2011}) but our approach could similarly be combined with more elaborate inverted indexes, such as the inverted multi-index~\cite{Babenko2015}. We divide the database into $K=65536$ inverted lists, and we scan $\mathit{ma}=64$ inverted lists to answer each query. In this context also, our approach achieves the same recall as 16-bit quantizers (Figure~\ref{fig:64ball}c). When inverted indexes are used, 16-bit quantizers lead to a very high increase in response time: 16-bit quantizers are more than 10 times slower than 8-bit quantizers (Figure~\ref{fig:64ball}c). This is because 16-bit quantizer cause a high increase in \emph{Tables} time and \emph{Scan} time. In the case of exhaustive search, the increase in \emph{Tables} time remains negligible (Section~\ref{sec:16bitimpact}, Table~\ref{tbl:impact}), despite the large increase. However, in the case of inverted indexes, the increase in \emph{Tables} becomes significant as multiple tables are computed (in our case $\mathit{ma}=64$ tables). By contrast, our approach achieves the same recall as 16-bit quantizers, but is only 1.1 to 1.5 times slower than 8-bit quantizers.

\subsection{Exhaustive search, 32-bit codes}
There has been a recent interest in very high compression levels, and in 32-bit codes \cite{Babenko2014AQ,Babenko2015TQ}. In this context, the use of 16-bit quantizers is particularly beneficial as they bring a higher increase in recall than for 64-bit codes. For space reasons, we do not include the full graphs for 32-bit codes but only report time and recall values for $r=100$ (Table \ref{tbl:32bcodes}). On the SIFT1M dataset and for 32-bit codes, 16-bit quantizers offer a 21\% increase in recall: the Recall@100 is 0.788 compared to 0.652 for 8-bit quantizers (OPQ, Table \ref{tbl:32bcodes}). On the TINY10M dataset, 16-bit quantizers offer a 24\% increase in recall: the Recall@100 is 0.347 compared to 0.278 for 8-bit quantizers (OPQ, Table \ref{tbl:32bcodes}). For 32-bit codes also, our approach provides the same recall as 16-bit quantizers, while offering a much lower response time. 

\begin{table}
  \centering
  \begin{threeparttable}
    \caption{Recall and response time (32-bit codes)\label{tbl:32bcodes}}
    \begin{tabularx}{\linewidth}{Xcc|cc}
      \toprule
      & \multicolumn{2}{c}{SIFT1M} & \multicolumn{2}{c}{TINY10M}\\
      \midrule
      & Time & R@100\tnote{1} & Time & R@100\tnote{2}\\
      \midrule
      \input{graphs/table_32b.booktabs.tex}
      \bottomrule
    \end{tabularx}
    \begin{tablenotes}
      \small
      \item \tnote{1} $r2=$10K\hspace{1em}\tnote{2} $r2=$120K
    \end{tablenotes}
  \end{threeparttable}
\end{table}

%% file: graphs/SIFT1M_r2.tex
\pgfplotsset{half width plot/.style={
  width=0.55*\linewidth,
  height=0.50*\linewidth,
}}

\begin{tikzpicture}[baseline]
\begin{semilogxaxis} [
  sfit1mr2 plot,
  half width plot,
  ylabel=Recall,
  xlabel=$r2$,
  name=recall graph,
  legend pos=south east,
  legend cell align=left,
  title={(a) Recall (r=100)}
]
  \addplot+
    table[x=r2, y=recall] {graphs/SIFT1M_r2_r100.dat};
  \addplot+
    table[x=r2, y=opq_recall] {graphs/SIFT1M_r2_r100.dat};
  \legend{PQ $4{\stimes}8,16$\\ OPQ $4{\stimes}8,16$\\}
\end{semilogxaxis}
\end{tikzpicture}
\begin{tikzpicture}[baseline, trim axis right]
\begin{semilogxaxis} [
  sfit1mr2 plot,
  half width plot,
  name=time graph,
  at={($(recall graph.east)+(3.7em,0cm)$)},
  anchor=west,
  xlabel=$r2$,
  ylabel=Time per query (ms),
  cycle list={blue},
  legend pos=north west,
  legend cell align=left,
  title=(b) Response time
]
  \addplot+[blue, mark=*]
    table[x=r2, y=query_us] {graphs/SIFT1M_r2_r100.dat};
  \addplot+[blue, mark=square*, dashed]
    table[x=r2, y=rerank_us] {graphs/SIFT1M_r2_r100.dat};
\legend{Total, Refine}
\end{semilogxaxis}
\end{tikzpicture}
%
%

%% file: graphs/time_recall_graphs.tex
\newcommand{\subTickFont}{\footnotesize}

\begin{tikzpicture}[baseline, trim axis left]
\begin{semilogxaxis} [
  r plot with dual axis,
  xticklabels={10\\{\subTickFont 9K},20\\{\subTickFont 9K},50\\{\subTickFont 9K},100\\{\subTickFont 9K},200\\{\subTickFont 12K},500\\{\subTickFont 12K},1K\\{\subTickFont 12K}},
  cycle list={black},
  legend pos=south east,
  ylabel=Recall,
  title = {(a) SIFT1M, no index}
]
  \addplot+ [solid, color=red, mark=*]
    table[x=r, y=pq8_recall] {graphs/SIFT1M_r_recall.dat};
  \addplot+ [solid, color=blue, mark=square*]
    table[x=r, y=pq16_recall] {graphs/SIFT1M_r_recall.dat};
  \addplot+ [solid, color=brown, mark=diamond*]
    table[x=r, y=pq816_recall] {graphs/SIFT1M_r_recall.dat};

  \addplot+ [dashed, color=red, mark=*]
    table[x=r, y=opq8_recall] {graphs/SIFT1M_r_recall.dat};
  \addplot+ [dashed, color=blue, mark=square*]
    table[x=r, y=opq16_recall] {graphs/SIFT1M_r_recall.dat};
  \addplot+ [dashed, color=brown, mark=diamond*]
    table[x=r, y=opq816_recall] {graphs/SIFT1M_r_recall.dat};

\end{semilogxaxis}
\end{tikzpicture}
\begin{tikzpicture}[baseline]
\begin{semilogxaxis} [
  r plot with dual axis,
  xticklabels={10\\{\subTickFont 60K},20\\{\subTickFont 60K},50\\{\subTickFont 70K},100\\{\subTickFont 100K},200\\{\subTickFont 120K},500\\{\subTickFont 150K},1K\\{\subTickFont 200K}},
  tick label style = {font=\small, align=center},
  cycle list={black},
  ylabel=Recall,
  title={(b) TINY10M, no index}
]
  \addplot+ [solid, color=red, mark=*]
    table[x=r, y=pq8_recall] {graphs/TINY10M_r_recall.dat};
  \addplot+ [solid, color=blue, mark=square*]
    table[x=r, y=pq16_recall] {graphs/TINY10M_r_recall.dat};
  \addplot+ [solid, color=brown, mark=diamond*]
    table[x=r, y=pq816_recall] {graphs/TINY10M_r_recall.dat};

  \addplot+ [dashed, color=red, mark=*]
    table[x=r, y=opq8_recall] {graphs/TINY10M_r_recall.dat};
  \addplot+ [dashed, color=blue, mark=square*]
    table[x=r, y=opq16_recall] {graphs/TINY10M_r_recall.dat};
  \addplot+ [dashed, color=brown, mark=diamond*]
    table[x=r, y=opq816_recall] {graphs/TINY10M_r_recall.dat};

\end{semilogxaxis}
\end{tikzpicture}
\begin{tikzpicture}[baseline, trim axis right]
\begin{semilogxaxis} [
  r plot with dual axis,
  xticklabels={10\\{\subTickFont 10K},20\\{\subTickFont 10K},50\\{\subTickFont 15K},100\\{\subTickFont 20K},200\\{\subTickFont 25K},500\\{\subTickFont 30K},1K\\{\subTickFont 30K}},
  tick label style = {font=\small, align=center},
  cycle list={black},
  legend pos=south east,
  ylabel=Recall,
  title={(c) SIFT1B, inverted index (K=65536)}
]
\addplot+ [solid, color=red, mark=*]
table[x=r, y=pq8_recall] {graphs/SIFT1B3M_r_recall.dat};
\addplot+ [solid, color=blue, mark=square*]
table[x=r, y=pq16_recall] {graphs/SIFT1B3M_r_recall.dat};
\addplot+ [solid, color=brown, mark=diamond*]
table[x=r, y=pq816_recall] {graphs/SIFT1B3M_r_recall.dat};

\addplot+ [dashed, color=red, mark=*]
table[x=r, y=opq8_recall] {graphs/SIFT1B3M_r_recall.dat};
\addplot+ [dashed, color=blue, mark=square*]
table[x=r, y=opq16_recall] {graphs/SIFT1B3M_r_recall.dat};
\addplot+ [dashed, color=brown, mark=diamond*]
table[x=r, y=opq816_recall] {graphs/SIFT1B3M_r_recall.dat};

\end{semilogxaxis}
\end{tikzpicture}

\begin{tikzpicture}[baseline, trim axis left]
\begin{semilogxaxis} [
  r plot with dual axis,
  xticklabels={10\\{\subTickFont 9K},20\\{\subTickFont 9K},50\\{\subTickFont 9K},100\\{\subTickFont 9K},200\\{\subTickFont 12K},500\\{\subTickFont 12K},1K\\{\subTickFont 12K}},
  cycle list={black},
  legend pos=north west,
  ylabel=Time per query (ms)
]
\addplot+ [solid, color=red, mark=*]
table[x=r, y=pq8_query_us] {graphs/SIFT1M_r_time.dat};
\addplot+ [solid, color=blue, mark=square*]
table[x=r, y=pq16_query_us] {graphs/SIFT1M_r_time.dat};
\addplot+ [solid, color=brown, mark=diamond*]
table[x=r, y=pq816_query_us] {graphs/SIFT1M_r_time.dat};

\addplot+ [dashed, color=red, mark=*]
table[x=r, y=opq8_query_us] {graphs/SIFT1M_r_time.dat};
\addplot+ [dashed, color=blue, mark=square*]
table[x=r, y=opq16_query_us] {graphs/SIFT1M_r_time.dat};
\addplot+ [dashed, color=brown, mark=diamond*]
table[x=r, y=opq816_query_us] {graphs/SIFT1M_r_time.dat};
\end{semilogxaxis}
\end{tikzpicture}
\begin{tikzpicture}[baseline]
\begin{semilogxaxis} [
  r plot with dual axis,
  xticklabels={10\\{\subTickFont 60K},20\\{\subTickFont 60K},50\\{\subTickFont 70K},100\\{\subTickFont 100K},200\\{\subTickFont 120K},500\\{\subTickFont 150K},1K\\{\subTickFont 200K}},
  tick label style = {font=\small, align=center},
  cycle list={black},
  legend pos=north west,
  ylabel=Time per query (ms),
  legend to name=pq-opq-64b-legend,
  legend columns=6
]
\addplot+ [solid, color=red, mark=*]
table[x=r, y=pq8_query_us] {graphs/TINY10M_r_time.dat};
\addplot+ [solid, color=blue, mark=square*]
table[x=r, y=pq16_query_us] {graphs/TINY10M_r_time.dat};
\addplot+ [solid, color=brown, mark=diamond*]
table[x=r, y=pq816_query_us] {graphs/TINY10M_r_time.dat};

\addplot+ [dashed, color=red, mark=*]
table[x=r, y=opq8_query_us] {graphs/TINY10M_r_time.dat};
\addplot+ [dashed, color=blue, mark=square*]
table[x=r, y=opq16_query_us] {graphs/TINY10M_r_time.dat};
\addplot+ [dashed, color=brown, mark=diamond*]
table[x=r, y=opq816_query_us] {graphs/TINY10M_r_time.dat};
\legend{PQ $8{\stimes}8$\\
  PQ $4{\stimes}16$\\
  PQ $4{\stimes}8,16$\\
  OPQ $8{\stimes}8$\\
  OPQ $4{\stimes}16$\\
  OPQ $4{\stimes}8,16$\\}

\end{semilogxaxis}
\end{tikzpicture}
\begin{tikzpicture}[baseline, trim axis right]
\begin{semilogxaxis} [
  r plot with dual axis,
  xticklabels={10\\{\subTickFont 10K},20\\{\subTickFont 10K},50\\{\subTickFont 15K},100\\{\subTickFont 20K},200\\{\subTickFont 25K},500\\{\subTickFont 30K},1K\\{\subTickFont 30K}},
  cycle list={black},
  legend pos=north west,
  ylabel=Time per query (ms)
]
\addplot+ [solid, color=red, mark=*]
table[x=r, y=pq8_query_us] {graphs/SIFT1B3M_r_time.dat};
\addplot+ [solid, color=blue, mark=square*]
table[x=r, y=pq16_query_us] {graphs/SIFT1B3M_r_time.dat};
\addplot+ [solid, color=brown, mark=diamond*]
table[x=r, y=pq816_query_us] {graphs/SIFT1B3M_r_time.dat};

\addplot+ [dashed, color=red, mark=*]
table[x=r, y=opq8_query_us] {graphs/SIFT1B3M_r_time.dat};
\addplot+ [dashed, color=blue, mark=square*]
table[x=r, y=opq16_query_us] {graphs/SIFT1B3M_r_time.dat};
\addplot+ [dashed, color=brown, mark=diamond*]
table[x=r, y=opq816_query_us] {graphs/SIFT1B3M_r_time.dat};

\end{semilogxaxis}
\end{tikzpicture}\\
\pgfplotslegendfromname{pq-opq-64b-legend}

%% file: graphs/table_32b.booktabs.tex
\pq{4}{8}   & \emph{1.36 ms} &       0.594  &  \emph{12.6 ms} & 0.126         \\
\pq{2}{16}  &       3.01 ms  & \emph{0.797} &        42.7 ms  & \emph{0.291}  \\
\pq{2}{8,16}& \emph{2.13 ms} & \emph{0.785} &  \emph{23.3 ms} & \emph{0.283}  \\\midrule
\opq{4}{8}   & \emph{1.31 ms} &       0.652 &  \emph{12.6 ms} & 0.278         \\
\opq{2}{16}  &       3.01 ms  & \emph{0.803}&        42.6 ms  & \emph{0.352}  \\
\opq{2}{8,16}& \emph{2.06 ms} & \emph{0.788}&  \emph{23.3 ms} & \emph{0.347}  \\
 

%% file: related.tex
\emph{Limitations.} As shown in Section~\ref{sec:eval}, our approach does not significantly increase response time while providing a substantial increase in recall. However, our approach still requires training 16-bit quantizers and encoding vectors with 16-bit quantizers, which can be costly. On our work workstation, encoding 1 million vectors with a $4{\stimes}16$ product quantizer takes 150 seconds (0.15 ms/vector), while this operation takes takes 0.91 seconds (0.00091 ms/vector) with a $8{\stimes}8$ product quantizer. Similarly, training the codebooks of a $4{\stimes}16$ product quantizer takes 4 minutes (50 k-means iterations, 100K SIFT descriptors), while training the codebooks of a $8{\stimes}8$ product quantizer takes 5 seconds. Nonetheless, it is unlikely that any of this disadvantages would have a practical impact. Codebooks are trained once and for all, therefore training time does not have much impact, as long as it remains tractable. Moreover, encoding a vector takes much less time than an ANN query (0.15 ms versus 2-20 ms), even with $4{\stimes}16$ product quantizers.

\emph{Comparison with Other Quantization Approaches.} Recently, compositional quantization models inspired by Product Quantization have been introduced to reduce quantization error. Among these models are Additive Quantization (AQ) \cite{Babenko2014AQ}, Tree Quantization (TQ) \cite{Babenko2015TQ} and Composite Quantization (CQ) \cite{Zhang2014}. Our approach compares favorably to these new quantization approaches as it does not significantly increase response time, and only moderately increases encoding time. On the contrary, AQ and TQ result in a more than twofold increase in ANN search time~\cite{Babenko2015TQ}. Moreover, AQ and TQ increase vector encoding time by multiple orders of magnitude: encoding a vector into a 64-bit code takes more than 200ms (1333 times more than our approach) for AQ, and 6ms for TQ (40 times more than our approach)~\cite{Babenko2015}. Even if encoding time is not as important as response time, such an increase makes AQ and TQ hardly tractable for large datasets. Thus, AQ and TQ have not been evaluated on datasets of more than 1 million vectors~\cite{Babenko2014AQ, Babenko2015TQ}. More importantly, our approach is orthogonal to the approach taken by AQ, TQ or CQ. Therefore, AQ, TQ or CQ could be used with 16-bit sub-quantizers, and associated derived codebooks, provided that vector encoding remains tractable. 

%

\emph{Use of Second-Order Residuals.} Like our approach, the ADC+R system~\cite{Jegou2011R} (also called IVFADC+R when it is combined with an inverted index), first builds a relatively large candidate set and then reranks it to obtain a final result set. The key idea of their approach is to encode a vector with a first $4{\stimes}8$ product quantizer, and to encode the residual error vector with a second $4{\stimes}8$ product quantizer. Unlike our approach, ADC+R cannot exploit 16-bit quantizers, and therefore does not benefit from the increase in accuracy they bring. The focus of ADC+R is on decreasing response time, at the expense of a slight decrease in accuracy. On the contrary, our approach increases accuracy (thanks to 16-bit quantizers) but comes at the expense of a slight increase in response time. 


\emph{Polysemous Codes.} Recently, the idea of polysemous codes \cite{Douze2016} that can be both interpreted as binary codes and PQ codes has been introduced. The binary codes allow a fast and approximate distance evaluation, which is then refined using the PQ codes (8-bit quantizers). The codes used in our approach can also be regarded as polysemous codes, that can be both interpreted as $4{\stimes}8$ PQ codes and $4{\stimes}16$ PQ codes. The major difference between our approach and the polysemous codes used in \cite{Douze2016} is that polysemous codes use binary codes for approximate distance evaluations, and PQ codes with 8-bit quantizer for precise distance evaluations. On the contrary, our approach uses PQ codes with 8-bit quantizers for approximate distance evaluation, and PQ codes with 16-bit quantizers for precise distance evaluation. In addition, we introduce a dynamic computation of lookup tables for 16-bit quantizers and the capped buckets data structure (Section~\ref{sec:budsearch}), which are not used in \cite{Douze2016}. The use of 16-bit quantizers allows our approach to achieve high recall: on the SIFT1B dataset, we achieve a Recall@100 of 0.850 in 5.2 ms. Unlike our approach, polysemous codes \cite{Douze2016} focus on moderate recall and extremely low response times: they achieve a Recall@100 of 0.332 in 0.27 ms on SIFT1B.

%% file: conclusion.tex
In this paper, we introduced the idea of derived codebooks, a novel approach that combines the speed of 8-bit quantizers with the accuracy of 16-bit quantizers. Our approach achieves the same recall as 16-bit quantizers for a response time close to the one of 8-bit quantizers. Moreover, our approach can be combined with inverted indexes, which is especially useful for large datasets. These results are achieved by building a set of derived codebooks ($2^{8}$ centroids each) from the codebooks of the 16-bit quantizers. The derived codebooks are used to build a candidate set as they allow fast and approximate distance evaluations. This candidate set is then refined using the 16-bit quantizers to obtain the final result set. 

As it provides a significant increase in recall without strongly impacting response time, our approach compares favorably with the state of the art. Because they incur a high increase in response time, 16-bit quantizers are generally believed to be intractable. With derived codebooks, we have shown that 16-bit quantizers can be time-efficient, thus opening new perspectives for ANN search based on product quantization.

Finally, our paper evaluates 8-bit derived codebooks for 16-bit quantizers but could be transposed to other settings such as 4-bit or 5-bit derived codebooks for 8-bit quantizers. Indeed, recent works~\cite{Andre2017,LeScouarnec2018c} have shown that SIMD-based implementations of 4,5 or 6-bit quantizers provide very low response time with good recall. Thus, they could be combined with the derived codebooks, introduced in this paper,to build \pq{8}{4,8} , \pq{8}{5,8}, \pq{8}{6,8}: The fast distance computation implemented in SIMD on the derived codebook can be used to prune the precise distance computation on the 8-bit codebooks. In that case, one may be willing to use symmetric distance computation (SDC) for the SIMD distance computation in order to avoid the overhead of distance table computation, similarly to polysemous codes~\cite{Douze2016} that also prune distance table computations.